\documentclass[10pt,journal,compsoc]{IEEEtran}
\ifCLASSOPTIONcompsoc
  \usepackage[nocompress]{cite}
\else
  \usepackage{cite}
\fi
\ifCLASSINFOpdf
\else
\fi
\usepackage{amssymb}
\usepackage{amsmath}
\usepackage{multirow}
\usepackage{booktabs}
\usepackage{graphicx}
\usepackage{ragged2e}
\usepackage{subfloat}
\usepackage{subfig}
\usepackage{color}

\usepackage[colorlinks=black,linkcolor=black]{hyperref}

\renewcommand{\justify}{\leftskip=0pt \rightskip=0pt plus 0cm}

\begin{document}
%
\title{DER-GCN: Dialogue and Event Relation-Aware Graph Convolutional Neural Network for Multimodal  Dialogue Emotion Recognition}
%
%
%
%

\author{Wei~Ai, Yuntao~Shou,
	Tao~Meng,
	Nan Yin, and~Keqin~Li,~\IEEEmembership{Fellow,~IEEE}
	\thanks{Corresponding Author: Tao Meng~(mengtao@hnan.edu.cn)}
\IEEEcompsocitemizethanks{\IEEEcompsocthanksitem W. Ai,~Y. Shou, and~T. Meng are with School of computer and Information Engineering, Central South University of Forestry and Technology, Hunan 410004,
	China. (mengtao@hnan.edu.cn,\ yuntaoshou@csuft.edu.cn,~aiwei@hnan.edu.cn)}
\IEEEcompsocitemizethanks{\IEEEcompsocthanksitem N. Yin is with Mohamed bin Zayed University of Artificial Intelligence, UAE. (nan.yin@mbzuai.ac.ae)}
\IEEEcompsocitemizethanks{
	\IEEEcompsocthanksitem K. L is with the Department of Computer Science, State University of New York, New Paltz, New York 12561, USA. (lik@newpaltz.edu)}
}

\IEEEtitleabstractindextext{%
\begin{abstract}
\justify
With the continuous development of deep learning (DL), the task of multimodal dialogue emotion recognition (MDER) has recently received extensive research attention, which is also an essential branch of DL. The MDER aims to identify the emotional information contained in different modalities, e.g., text, video, and audio, in different dialogue scenes. However, existing research has focused on modeling contextual semantic information and dialogue relations between speakers while ignoring the impact of event relations on emotion. To tackle the above issues, we propose a novel Dialogue and Event Relation-Aware Graph Convolutional Neural Network for Multimodal Emotion Recognition (DER-GCN) method. It models dialogue relations between speakers and captures latent event relations information. Specifically, we construct a weighted multi-relationship graph to simultaneously capture the dependencies between speakers and event relations in a dialogue. Moreover, we also introduce a Self-Supervised Masked Graph Autoencoder (SMGAE) to improve the fusion representation ability of features and structures. Next, we design a new Multiple Information Transformer (MIT) to capture the correlation between different relations, which can provide a better fuse of the multivariate information between relations. Finally, we propose a loss optimization strategy based on contrastive learning to enhance the representation learning ability of minority class features. We conduct extensive experiments on the IEMOCAP and MELD benchmark datasets, which verify the effectiveness of the DER-GCN model. The results demonstrate that our model significantly improves both the average accuracy and the f1 value of emotion recognition.
\end{abstract}

}

\maketitle

\IEEEdisplaynontitleabstractindextext

%
\IEEEpeerreviewmaketitle

\IEEEraisesectionheading{\section{Introduction}\label{sec:introduction}}

%
%
%
%
\subsection{Motivation}
\IEEEPARstart{T}{he} task of Multimodal Dialogue Emotion Recognition (MDER) is to identify the emotional changes of speakers in different modalities, such as text, video, and audio. In recent decades, due to the application of MDER in some emerging application scenarios, for instance, the recognition of negative emotions has attracted research attention in social media such as Meta and Weibo \cite{khare2020time}, \cite{10113198}, \cite{10314020}, the intelligent recommendation system for online shopping \cite{9219228}, and chat robots \cite{9839579}, etc. Furthermore, when shopping online, the APP will recommend the most interesting products according to the user's preferences \cite{yin2023coco}, \cite{10.1145/3503161.3548012}.

\begin{figure}
	\centering
	\includegraphics[width=1\linewidth]{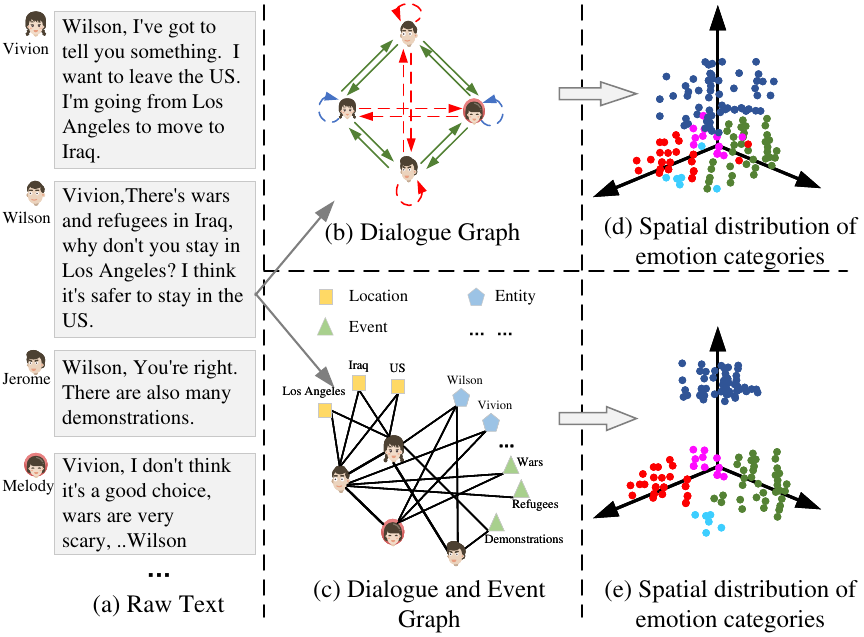}
	\caption{An illustrative example of the impact of event relationships on the spatial distribution of emotion categories. (a) Raw dialogue text with four speakers. (b) A graph of dialogue relationships composed of emotional interactions between speakers. (c) The emotional interaction graph is composed of dialogue and event relationships. (d) Spatial distribution of emotion categories in graphs composed of dialogue relations. (e) Spatial distribution of emotion categories in graphs composed of dialogue and event relationship.}
	\label{fig: event}
\end{figure}

However, MDER is more challenging than sentence-level emotion recognition or unimodal emotion recognition tasks because sentiment changes are generally determined by a series of meaningful internal and external factors \cite{shou2022conversational}, \cite{shou2022object}. Specifically, in the dialogue process, the speaker's emotion is affected not only by internal factors composed of contextual information but also by external factors composed of dialogue and event relationships (e.g., entity, location, keywords, etc.). For example, when speakers talk about a sensitive topic on social media, they often express their emotions more implicitly and suggestively \cite{meng2023multi}. While the use of events to strengthen the dialogue between speakers can compensate for the lack of noticeable semantic features. Therefore, how to comprehensively consider the influence of internal and external factors on emotion recognition is still a problem to be solved. In addition, in MDER, due to the high cost of labeling, the data distribution exhibits a long-tailed state \cite{ying2021prediction}, \cite{meng2023multi}, \cite{meng2023deep}. It leads to the model being less effective at identifying the minority class emotion.

The current mainstream MDER methods are used Recurrent Neural Networks (RNNs), \cite{Baddar_Ro_2019}, Transformers \cite{lian2021ctnet}, and Graph Neural Networks (GNNs) \cite{9540311} to model the semantic information of context and dialogue relationship between speakers, respectively. Although RNN-based methods have achieved good results in emotion recognition tasks based on contextual semantic modeling \cite{shou2023graph}, \cite{shou2023czl}, it ignores the influence of external factors (e.g., dialogue relations and event relations). To better integrate contextual semantic information, Transformer-based methods are applied, but it still ignores the influence of external factors on emotion recognition. To consider the influence of internal and external factors on emotion recognition, many researchers have begun to adopt GNN to model MDER \cite{shou2023comprehensive}. Although the GNN-based method considers the dialogue relationship, it still ignores the influence of the event relationship on MDER. However, the event relationship also greatly influences the speaker's emotion, and the speaker usually shows the same emotion when discussing the same event. Therefore, modeling the event relations in the dialogue is beneficial to obtain a better spatial distribution of emotion categories. As shown in Fig. 1 as an example, Fig. 1(b) is a graph that only considers the dialogue relationship between speakers, and its emotion categories have overlapping areas in the spatial distribution, as shown in Fig. 1 (d). Fig. 1(c) is a graph that comprehensively considers the interaction relationship and event relationship between speakers. Its spatial distribution of emotion categories has better discrimination, as shown in Fig. 1(e). Hence, it is necessary to take the event relationship as the starting point of the MDER architecture design.

To tackle the above problem, we propose a novel Dialogue and Event Relation-Aware Graph Convolutional Neural Network for Multimodal Emotion Recognition architecture, namely DER-GCN. DER-GCN mainly includes six modules: data preprocessing, feature extraction and fusion, masked graph representation learning, multi-relational information aggregation, balanced sampling strategy, and emotion classification. Firstly, we use RoBERTa \cite{Liu2019RoBERTaAR}, 3D-CNN \cite{8671459}, and Bi-LSTM-based Encoder \cite{NEURIPS2018_6832a7b2} to obtain embedding representations for three modalities: text, video, and audio. Secondly, we use a bidirectional gated recurrent unit (Bi-GRU) for feature extraction and Doc2EDAG \cite{zheng2021revisiting} for event extraction to strengthen the dialogue relationship between speakers. Then, we design a novel cross-modal feature fusion method to learn complementary semantic information between different modalities. Specifically, we use cross-modal attention to learn the differences between the semantic information of different modes. The average pooling operation is used to learn the global information of each mode to guide the inter-modality and intra-modality information aggregation, respectively. Thirdly, we design a self-supervised mask graph autoencoder (SMGAE) to model the correlation between dialogues and events. Unlike the previous works \cite{10.1145/3534678.3539321}, which only perform mask reconstruction on nodes in the graph, SMGAE performs mask reconstruction on some nodes and edges simultaneously. Fourthly, we design the Multiple Information Transformer (MIT) to better fuse the multivariate information between relations and capture the correlation between different relations. MIT is paid attention mechanism to filter unimportant relational information, which fuses to obtain better embedding representations. Fifthly, we propose a loss optimization function based on contrastive learning to alleviate the long-tail effect in MDER, which balances the proportion of each emotion category during model training. Finally, we have used an emotion classifier constructed from a multilayer perceptron (MLP) to output the final sentiment category.

\subsection{Our Contributions}
Hence, MDER should consider the dialogue between speakers and the event relationship in the dialogue as the starting point of model design. Inspired by the analysis above, we present a novel Dialogue and Event Relation-Aware Graph Convolutional Neural Network for Multimodal Emotion Recognition (DER-GCN) to learn better emotion feature embedding. The main contributions of this paper are summarized as follows:

\begin{itemize}
	\item A novel dialogue and event relation-aware emotion representation learning architecture is present and named DER-GCN. DER-GCN can achieve cross-modal feature fusion, solve the imbalanced data distribution problem, and learn more discriminative emotion class boundaries.
	
	
	\item  A novel self-supervised graph representation learning framework, named SMGAE, is presented. SMGAE enhances the feature representation capability of nodes and optimizes the structural representation of graphs, which has a stronger anti-noise ability.
	
	\item	A new Weighted Relation-aware Multi-subgraph Information Aggregation method is implemented and named MIT. MIT is used to learn the importance of different relations in information aggregation to fuse to obtain more discriminative feature embeddings.
	
	
	\item  Finally, extensive experiments are performed on two popular benchmark datasets, MELD and IEMOCAP, which demonstrate that DER-GCN outperforms existing comparative algorithms in weight accuracy and  f1-value for multimodal emotion recognition.
\end{itemize}

The remainder of this paper is organized as follows. Section 2 reviews the related work of the predecessors. Section 3 defines the problem and briefly describes how the data is preprocessed. Section 4 illustrates the structural design of DER-GCN. The datasets, evaluation metrics, and comparison algorithms are described in Section 5. Section 6 shows the experimental results. Finally, we present experimental conclusions and future work in Section 7.

\section{Related Work}

\subsection{Emotion Recognition in Conversation}

Multimodal Dialogue Emotion Recognition (MDER) is an interdisciplinary research field that has attracted extensive attention from researchers in cognitive science, psychology, etc. Existing MDER research mainly includes emotion recognition based on Recurrent Neural Network (RNN) \cite{latif2021survey}, emotion recognition based on Graph Neural Network (GNN) \cite{kong2022causal}, and emotion recognition based on Transformer \cite{lian2021ctnet}. RNNs mainly extract contextual semantic information by modeling long-range contextual dependencies. GNNs model, the dynamic interaction process of dialogue, mainly relies on the graph structure's inherent properties to model the dependencies between speakers. The Transformer mainly uses the attention mechanism to achieve cross-modal feature fusion to capture the different semantic information between modalities.

In the RNN-based multimodal emotion recognition research, Wang et al. \cite{9054629} conducted Dual-Sequence LSTM (DS-LSTM), which uses a dual-stream LSTM to extract contextual features in the Mei-Frequency map simultaneously. DS-LSTM comprehensively considers the context features of different times and frequencies and achieves a better emotion recognition effect. However, DS-LSTM cannot achieve feature fusion between modalities. Li et al. \cite{li2020exploring} created attention-based bidirectional LSTM RNNs (A-BiLSTM RNNs). This method combines the self-attention mechanism and LSTM to learn multimodal features with a time dimension. It carries out decision-level information fusion on the obtained multimodal features to realize emotion recognition. Although RNN-based methods have achieved good results in emotion recognition tasks based on contextual semantic modeling, it ignores the influence of external factors (e.g., dialogue relations and event relations).

In Transformer-based multimodal emotion recognition research, Huang et al. \cite{9053762} employed Multimodal Transformer Fusion (MTF), which uses a multi-head attention mechanism to obtain intermediate feature representations of multimodal emotions. Then, a self-attention mechanism is utilized to capture long-lived dependencies in context. MTF significantly outperforms other feature fusion methods. Transformer-based methods can extract richer contextual semantic information, but it still ignores the influence of external factors on emotion recognition.

In GNN-based multimodal emotion recognition research, Sheng et al. \cite{sheng-etal-2020-summarize} performed Summarization and Aggregation Graph Inference Network (SumAggGIN), which captures distinguishable fine-grained features between phrases by building a heterogeneous graph neural network. SumAggGIN achieves sentiment prediction related to dialogue topics. Although the GNN-based method considers the dialogue relationship, it still ignores the influence of the event relationship on MDER.

{\subsection{Transformers for Dialogue Generation}}
{In recent years, the task of dialogue generation has also begun to receive extensive attention. For example, dialogue generation technology can be used in healthcare to help patients access health information. Huang et al. proposed Persona-Adaptive Attention (PAA), which uses a dynamic mask attention mechanism to adaptively reduce redundant information in context information. Zheng et al. proposed a pre-trained personalized dialogue model, which uses a large-scale pre-trained model to initialize model weights, and introduces attention in the decoder to dynamically extract context information and role information.  Zeng et al. introduced a condition-aware Transformer to generate probability deviations for words in different positions.}

\subsection{Masked Self-Supervised Graph Learning}
Masked self-supervised graph representation learning, which can automatically learn deeper feature representations from raw data without using a large amount of labeled data, has been used by more and more researchers. The current mainstream research on mask self-supervised graph representation learning focuses on mask and data reconstruction at the node and edge levels.

In node-level mask-based self-supervised learning, Liu et al. \cite{liu2021spatiotemporal} performed a spatiotemporal graph neural network (STG-Net), which masks graph nodes based on an edge weighting strategy. GCN is used to reconstruct contextual features to obtain a better data representation. Wang et al. \cite{10.1145/3447548.3467415} created HeCo, which learns high-level embedding representations of nodes by using a view masking mechanism. In addition, HeCo introduces a contrastive learning strategy, which can further improve the model's ability to learn feature representations. However, HeCo needs to design meta-paths manually. The above methods only consider feature mask reconstruction and ignore structure mask reconstruction.

In edge-level mask-based self-supervised learning, Pan et al. \cite{10.1145/3447548.3467415} conducted adversarial graph embedding (AGE), which reconstructs the topology of a graph by using an adversarial regularized graph auto-encoder (ARGA) and an adversarial regularized variational graph auto-encoder (ARVGA). AGE is trained in a self-supervised manner to learn the underlying distribution law of the data. However, AGE may lose some useful semantic information. The above methods only consider structure mask reconstruction and ignore feature mask reconstruction.

\subsection{Balanced Optimization Based on Contrastive Learning}

The datasets in multimodal dialogue emotion recognition suffer from data imbalance, which makes the cross-entropy loss function widely used for classification no longer applicable. However, contrastive learning can learn distinguishable class boundary information between different classes by continuously narrowing the gap between positive samples. It continuously widens the gap between positive and negative samples \cite{kalantidis2020hard}. Therefore, contrastive learning is often used to solve the data imbalance problem in practical problems.

Cai et al. \cite{cai2022heterogeneous} applied a Heterogeneous Graph Contrastive Learning Network (HGCL), which obtains the embedded representation of each node by maximizing the interaction information between local graph nodes and the global representation of the full graph nodes. HGCL can learn better class boundary information from multivariate heterogeneous data. Peng et al. \cite{peng2022open} proposed supervised contrastive learning (SCL) to compare the input samples with other instances and input samples with negative samples, which were generated by the Soft Brownnian Offset sampling method to enhance feature representation capability. SCL can effectively alleviate the problem of imbalanced data distribution by continuously expanding the difference between positive and negative samples. However, SCL may suffer from overfitting because the feature representation is too strong.

\section{Preliminary Information}
In this section, we will define the multimodal emotion recognition task and briefly introduce the preprocessing methods for the three modalities of text, audio and video in the multimodal emotion dataset. Their processing procedures are as follows: (1) Word Embedding: To obtain word vectors with rich semantic information, we will use the RoBERTa model \cite{Liu2019RoBERTaAR} to obtain the vector representation of each word. (2) Visual Feature Extraction: To capture the features of the speaker's facial expression changes and gesture changes in each frame of the video, we will use the 3D-CNN model \cite{9363624} for feature extraction. (3) Audio Feature Extraction: To capture the speech features that can distinguish different speakers, we would use the structure of the Encoder to extract the feature of the sound signal. In addition, to capture the semantic information of capturing the topic events discussed by the speaker during the dialogue, we also perform event extraction on the text.

\subsection{Problem Definition}
For the task of MDER, since the number of speakers $N(N \geq 2)$ participating in the dialogue is not fixed, we assume that $N$ speakers involved in a conversation and are represented as $P=\{P_1,P_2,\ldots,P_N\}$, respectively. During the dialogue, a series of utterances from the speaker are arranged in chronological order, which can be expressed as $U=\{u_1,u_2,\ldots,u_T\}$. Where $T$ represents the total number of utterances, each of utterance has three modalities, i.e., text (t), audio (a), and visual (v). The task of this paper is to predict the speaker's emotion category at the current moment $q$ based on the speaker's words, voice, and his expressions. The emotion prediction task is defined as follows:
\begin{equation}
	e_i=\operatorname{prediction}\left(\left\{u_1, u_2, \ldots, u_i\right\}\right), i \in[q-K, q-1]
\end{equation}
where $e_i$ represents the emotion of the $i$-th sentence, and $K$ represents the window size of the historical context.

\begin{figure*}
	\centering
	\includegraphics[width=0.95\linewidth]{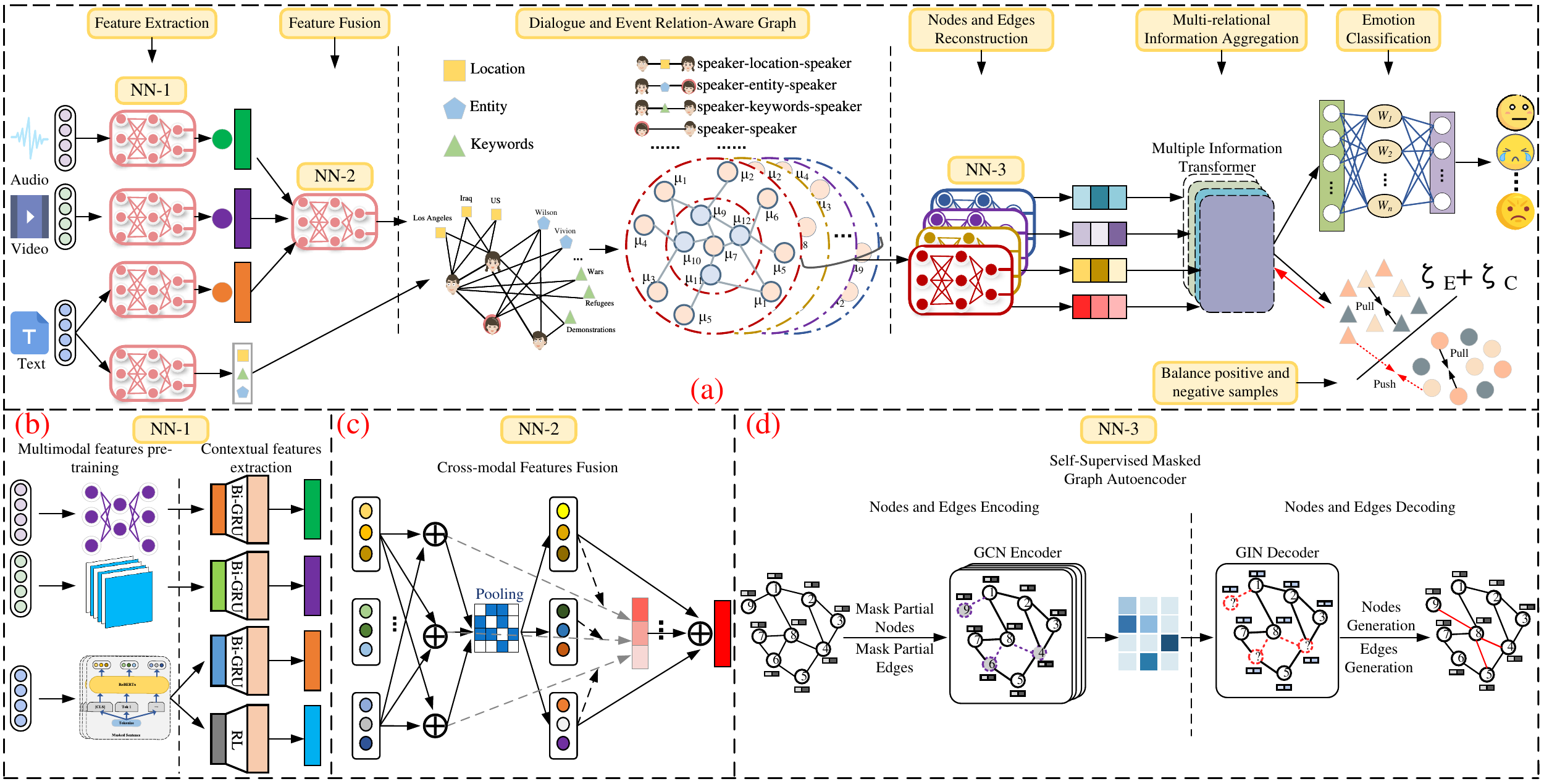}
	\caption{(a) The overall process framework of DER-GCN: It first preprocesses multimodal data to obtain encoded feature embeddings via NN-1. Second, it uses NN-2 to achieve cross-modal feature fusion. Third, it constructs a weighted multi-relational dialogue and event relation-aware graph through the fused feature vectors. Fourth, node and edge features are reconstructed via NN-3. Fifth, the fused multi-relational information feature vectors are obtained through the Multiple Information Transformer, and a loss optimization strategy based on contrastive learning is used to solve the data imbalance problem. Finally it uses the emotion classifier to get the final emotion label. (b) NN-1: multimodal feature encoder. (c) NN-2: cross-modal feature aggregator. (d) NN-3: self-supervised masked graph encoder.}
	\label{fig:graphmae}
\end{figure*}

\section{Methodology}
\subsection{The Design of the DER-GCN Structure}
In this section, we illustrate the six components that makeup DER-GCN, as shown in Fig. 2. The structure of DER-GCN is as follows: (1) Sequence modeling and cross-modal feature fusion: For the input text, video, and audio modal features, DER-GCN inputs them into the bidirectional gated recurrent unit (Bi-GRU) to extract contextual semantic information. Furthermore, to capture the regions with the strongest emotional features among the three modalities, we design a cross-modal attention mechanism for feature extraction and fusion of complementary semantic information. (2) Multi-relational emotional interaction graph: Unlike current mainstream algorithms that only use graph convolutional neural networks (GCN) to model the interaction between speakers, we construct a multi-relational graph neural network that includes events and speakers, thereby enhancing the feature representation capability of the model. (3) Intra-relational Masked Graph Autoencoder: To improve the fusion representation ability of node features and edge structures in GCN, we designed a Masked Graph Autoencoder (MGAE). MGAE improves the representation ability of GCN by random masking and reconstruction of nodes and edges and alleviates the problem of class distribution imbalance. 4) Information aggregation between relations: To guide DER-GCN better to perform information aggregation of multi-relational graph neural networks, we design a multi-relational information fusion Transformer, which can effectively fuse the semantic information in the subgraphs composed of different relationships and learn better-embedded representation. (5) Contrastive Learning: The commonly used benchmark datasets in the field of multimodal emotion recognition have the problem of unbalanced class distribution. We introduce a contrastive learning mechanism to learn more discriminative class boundary information. (6) Emotion classifier: To make DER-GCN provide more gradient information in the backpropagation process and promote the model to be fully trained during emotion classification, we construct a linear layer with residual connections as the emotional classifier of DER-GCN.

\subsubsection{Sequence Modeling and Cross-modal Feature Fusion}

The emotional change of the speaker at the current time $t$ is not only related to the utterance at the $t$-th time but also to the contextual utterances before the $t-1$ time and after the $t+1$ time. How capture the contextual semantic information contained in the three modalities of video, audio and text is a challenging task. In this paper, we use Bi-GRU to model the long-term dependencies of the three modalities so that the model can more accurately understand the emotional changes of the speaker at the current moment $t$. The formula for GRU is defined as follows:

\begin{equation}
	\begin{gathered}
		z_{t}^{\gamma}=\operatorname{sigmoid}\left(W_{z}^{\gamma} \cdot\left[h_{t-1}^{\gamma}, x_{t}^{\gamma}\right]\right) \\
		r_{t}^{\gamma}=\operatorname{sigmoid}\left(W_{r}^{\gamma} \cdot\left[h_{t-1}^{\gamma}, x_{t}^{\gamma}\right]\right) \\
		\tilde{h}_{t}^{\gamma}=\tanh \left(W_{\widetilde{h}_{t}}^{\gamma} \cdot\left[r_{t}^{\gamma} \odot h_{t-1}^{\gamma}\right]\right) \\
		h_{t}^{\gamma}=\left(1-z_{t}^{\gamma}\right) \odot h_{t-1}^{\gamma}+z_{t}^{\gamma} \odot \tilde{h}_{t}^{\gamma}
	\end{gathered}
\end{equation}
where, $z_t$ represents the update gate, which is used to select the context information that needs to be retained at the current time $t$ to update the state of the hidden layer at the $t-1$-th time. $r_t$ represents the reset gate, which is used to forget the unimportant contextual information in the conversation at the current moment $t$. $x_t$, $h_t$ represent the input unimodal feature vector and the hidden layer for storing contextual information. $\tilde{h}_t$ represents the candidate's hidden layer state. $W_z$, $W_r$, $W_{\tilde{h}_t }$ are parameters that can be learned in GRU. $\gamma$ represents text, video or audio. $\odot$ means Hadamard product.

Bi-GRU contains contextual semantic information extracted from forward and reverse. The formula is defined as follows:
\begin{equation}
	\begin{gathered}
		\delta_{t}=\left[\stackrel{\rightarrow}{h}_{t}^{\gamma}: \stackrel{\leftarrow}{h}_{t}^{\gamma}\right] \\
		\psi^{\gamma}=\operatorname{concat}\left(\left[\delta_{1}, \delta_{2}, \ldots, \delta_{T}\right]\right)
	\end{gathered}
\end{equation}
where, $\stackrel{\rightarrow}{h}_{t}^{\gamma}$ is the contextual semantic information extracted in the forward direction, $\stackrel{\leftarrow}{h}_{t}^{\gamma}$ is the contextual information extracted in the reverse direction, and $\psi$ is composed of all the contextual information at the previous $T$ moments.

To realize the information interaction and fusion among the three modalities, we propose a cross-modal attention mechanism, which is used to exploit the interaction between modalities in a more fine-grained manner to improve the semantic understanding ability of the model.

First, we normalize the hidden layer feature vectors of the three modalities obtained after Bi-GRU processing. The formula is defined as follows:
\begin{equation}
	\mathcal{H}_{i j}^{\gamma}=\frac{\exp \left(\varepsilon^{\gamma} \psi_{i j}^{\gamma}\right)}{\sum_{m=1}^{n} \exp \left(\varepsilon^{\gamma} \psi_{k j}^{\gamma}\right)}
\end{equation}
where, $\mathcal{\varepsilon}^{\gamma}=\frac{1}{\sqrt{d^{\gamma}}}$ is the scaling factors of the three modalities. $n$ represents the dimension of the modality.

\begin{figure*}
	\centering
	\includegraphics[width=0.95\linewidth]{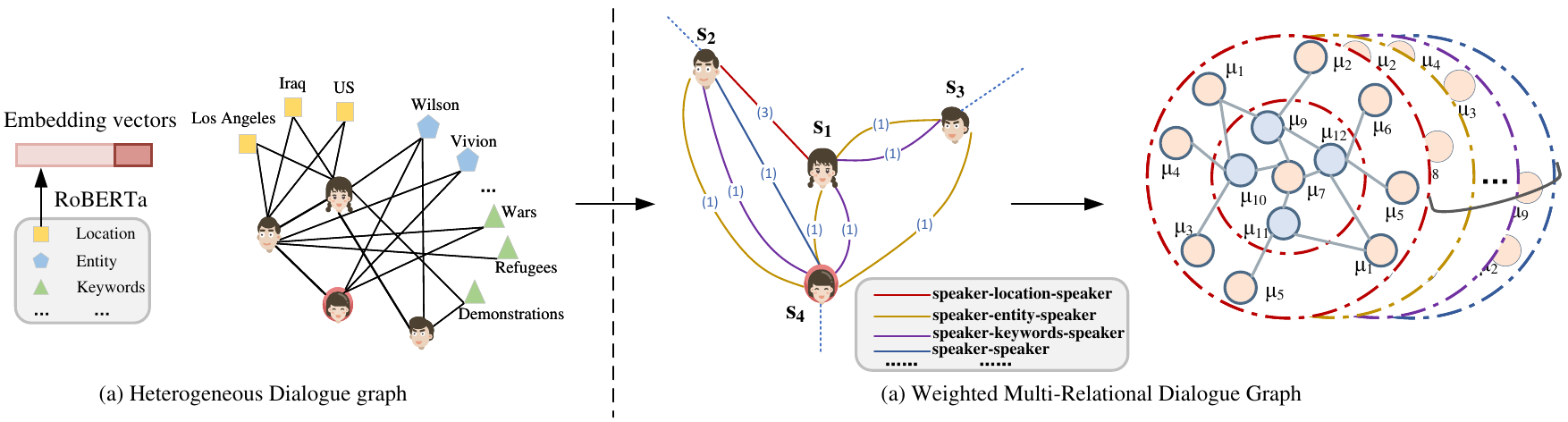}
	\caption{(a) Heterogeneous dialogue graph composed of dialogue relations and event relations. (b) We split the heterogeneous graph to construct a weighted multi-relational dialogue graph.}
	\label{fig:jiaquan}
\end{figure*}

Then, to better preserve the semantic information of the three modalities, we perform an average pooling operation on $H_{ij}^\gamma$, and the formula is defined as follows:
\begin{equation}
	\xi_{t}, \xi_{a}, \xi_{v}=f_{\text {pooling }}\left(\mathcal{H}^{T}, \mathcal{H}^{A}, \mathcal{H}^{U}\right)
\end{equation}
where $f_{pooling} (\cdot)$ represents the average pooling operation.

Next, we perform a fusion operation on the three modal features and use the tanh activation function to obtain their weights. The formula is defined as follows:
\begin{equation}
	\begin{aligned}
		\omega_{i}^{t} &=W^{T} \tanh \left(\lambda_{v} \mathcal{H}_{v}+\lambda_{a} \mathcal{H}_{a}+\lambda_{t} \xi_{t}+b_{t}\right) \\
		\omega_{i}^{a} &=W^{A} \tanh \left(\lambda_{t} \mathcal{H}_{t}+\lambda_{v} \mathcal{H}_{v}+\lambda_{a} \xi_{a}+b_{a}\right) \\
		\omega_{i}^{v} &=W^{V} \tanh \left(\lambda_{t} \mathcal{H}_{t}+\lambda_{a} \mathcal{H}_{a}+\lambda_{v} \xi_{v}+b_{v}\right)
	\end{aligned}
\end{equation}
where, $W^T$, $W^A$, $W^V$, $\lambda_a$, $\lambda_v$, $\lambda_t$, $b_t$, $b_a$, $b_v$ are the network parameters that can be learned in the model. According to the above formula, we can get the normalized attention weight $\widetilde{\omega}_{i}=\left\{\widetilde{\omega}^{t}, \widetilde{\omega}^{a}, \widetilde{\omega}^{v}\right\}$. The formula is defined as follows:
\begin{equation}
	\widetilde{\omega}_{i}=\frac{\omega_{i}}{\sum_{m=1}^{k} \omega_{k}}
\end{equation}

Finally, we obtain the feature representation $\xi$ after the fusion of the three modalities according to the attention weight. The formula is defined as follows:
\begin{equation}
	\xi=\xi_{t} * \widetilde{\omega}_{i}^{t}+\xi_{a} * \widetilde{\omega}_{i}^{a}+\xi_{v} * \widetilde{\omega}_{i}^{v}
\end{equation}

\subsubsection{Weighted Multi-relational Affective Interaction Graph}
As shown in Fig. 3, we build a multi-relational affective interaction graph that includes the relationships between speakers and heterogeneous elements extracted from events. To capture the heterogeneous information contained in different relations, we construct a weighted multi-relational affective interaction graph $G=\left\{V, \aleph, W,\left\{\Re_{r}^{\omega}\right\}_{r=1}^{R}\right\}$ to associate the relationship between nodes edges. The node set $V$ in the multi-relational emotional interaction graph is a series of fused multimodal feature vectors. The edge $e_{ij}\in \aleph$ is composed of speaker relation or event relation between $v_i$ and $v_j$. $\omega_{ij}\in W$ is the weight of the edge $e_{ij}$. $r\in \Re$ is an edge relation.

The formula for the edge $e_{ij}^{r_E} $of different relations composed of events is defined as follows:
\begin{equation}
	e_{i j}^{r_{E}}=\min \left\{\left[A_{r_{E}} \cdot A_{r_{E}}^{T}\right]_{i j}, 1\right\}
\end{equation}
where, $A_{r_E}$ represents the adjacency matrix of the multi-relationship graph, where its rows represent all event nodes, and its columns represent event nodes belonging to relation $r_E$. $A_{r_E}^T$ is the transposition of the matrix $A_{r_E}$. To capture the difference between different edges under the same relationship, we define the weight of the edge $e_{ij}^{r_E}$ as follows:
\begin{equation}
	\omega_{i j}^{r_{E}}=\left[A_{r_{E}} \cdot A_{r_{E}}^{T}\right]_{i j}
\end{equation}

For the edge $e_{ij}^{r_S}$ composed of the relationship between the speakers, if there is a dialogue between the speakers, we connect an edge for them. Otherwise, no edge is established. For the edge weight $\omega_{ij}^{r_S}$ of edge $e_{ij}^{r_S}$, we use the similarity attention mechanism to assign weights to it. First, we use two linear layers to compute the similarity between nodes in the graph. The formula is defined as follows:
\begin{equation}
	\rho_{i j}^{r_{S}}=W_{\varpi_{1}}^{r_{S}}\left(\operatorname{ReLU}\left(W_{\varpi_{2}}^{r_{S}}\left[\xi_{i}^{r_{S}} \oplus \xi_{j}^{r_{S}} \otimes \mathfrak{I}_{i j}\right]+b_{2}\right)+b_{1}\right)
\end{equation}
where, $W_{\varpi_{1}}^{r_{S}}$,$W_{\varpi_{2}}^{r_{S}}$ are the learnable parameters in the linear layer. $b_1$, $b_2$ are the biases of the linear layers. $\oplus$ means splicing, $\otimes$ means dot multiplication. $\mathfrak{I}_{i j}\in \{0,1\}$.

Then, we use the attention mechanism to get the weight of each edge, and the formula is defined as:
\begin{equation}
	\omega_{i j}^{r_{S}}=\operatorname{softmax}\left(\rho_{i j}^{r_{S}}\right)=\frac{\exp \left(\rho_{i j}^{r_{S}}\right)}{\sum_{n \in \mathcal{M}_{i}} \exp \left(\rho_{i m}^{r_{S}}\right)}
\end{equation}
where $\mathcal{M}_{i}$ is the set of neighbor nodes of node $i$. The larger $\omega_{ij}$, the higher the correlation between nodes.

\subsubsection{Self-Supervised Masked Graph Autoencoder}
To improve the joint representation ability of features and structures of graph neural networks, we propose a self-supervised masked graph autoencoder (SMGAE), which learns better feature embedding representation by randomly masking and reconstructing the nodes and edges in the graph. Unlike recent studies that only reconstruct features or structures, we reconstruct both features and structures to improve the generalization performance of the model.

First, we sample some nodes and edges in the graph and use the mask token to mask the node's feature vector and edge weights. The node feature formula after masking is defined as follows:
\begin{equation}
	\tilde{\xi}_{i}= \begin{cases}\xi_{[M]} & v_{i} \in V_{M} \\ \xi_{i} & v_{i} \notin V_{M}\end{cases}
\end{equation}
where, $V_M$ represents the masked node-set, and $\xi_{[M]}$ is the masked multimodal feature vector.

The formula for the masked edge is defined as follows:
\begin{equation}
	\tilde{e}_{i j}=\left\{\begin{array}{cl}
		e_{i j}^{[M]} & \aleph_{i} \in \varphi_{M} \\
		e_{i j} & \aleph_{i} \notin \varphi_{M}
	\end{array}\right.
\end{equation}
where, $\varphi_M$ represents the masked edge set, and $e_{ij}^{[M]}$ represents the masked edge.

The goal of SMGAE is to reconstruct the masked node features and adjacency matrix $A$ by using a small number of node features and edge weights. This paper is adopted a graph convolutional neural network (GCN) as our Encoder to aggregate information. The formula is defined as:
\begin{equation}
	\begin{aligned}
		&	p_{\vartheta}\left(\xi_{i}, e_{i} \mid \hat{\xi}_{i}, \hat{e}_{i}\right) \\ &=\sum_{M} p_{\vartheta}\left(\xi_{i}, e_{i}^{[M]} \mid e_{i}^{\sim[M]}, \hat{\xi}_{i}, \hat{e}_{i}\right) \cdot p_{\vartheta}\left(e_{i}^{[M]} \mid \hat{\xi}_{i}, \hat{e}_{i}\right) \\
		&=\mathbb{E}_{[M]}\left[p_{\vartheta}\left(\xi_{i}, e_{i}^{\sim[M]} \mid e_{i}^{[M]}, \hat{\xi}_{i}, \hat{e}_{i}\right)\right] \\
		&=\mathbb{E}_{[M]}\left[p_{\vartheta}\left(\xi_{i} \mid e_{i}^{[M]}, \hat{\xi}_{i}, \hat{e}_{i}\right) \cdot p_{\vartheta}\left(e_{i}^{\sim[M]} \mid e_{i}^{[M]}, \hat{\xi}_{i}, \hat{e}_{i}\right)\right]
	\end{aligned}
\end{equation}
where, $p_{\vartheta}\left(\xi_{i} \mid e_{i}^{[M]}, \hat{\xi}_{i}, \hat{e}_{i}\right)$ is the expected value of the generated node feature, and $p_{\vartheta}\left(e_{i}^{\sim[M]} \mid e_{i}^{[M]}, \hat{\xi}_{i}, \hat{e}_{i}\right)$ is the expected value of the generated edge. $e_{i}^{[M]}$ is the unmasked edge. $\hat{\xi}_{i}$, $\hat{e}_{i}$ represent the node features and edges generated by encoding, respectively.

In this paper, we will use a graph convolutional neural network (GCN) as our encoder to aggregate information, the formula is defined as:
\begin{equation}
	\begin{aligned}
	I_{i}^{(t)}=&\operatorname{ReLU}\Bigg(\sum_{k \in \aleph_{i}^{r}} \sum_{r \in \Re} \sum_{j \in \mathcal{M}_{i}^{r}}\Big(\frac{w_{i j}^{r}}{c_{i, r}} W_{r}^{(t)} I_{j}^{(t-1)}\\ &+w_{i i}^{r} W_{\zeta}^{(t)} I_{i}^{(t-1)}\Big) \cdot \tilde{e}_{i k}\Bigg)
	\end{aligned}
\end{equation}
where, $I_i^{(t)}$ is the feature vector representation of node $i$ at time $t$. $\aleph_{i}^{r}$ represents the edge set of node $i$ under the edge relation $r \in\{\Re\}_{r=1}^{R}$. $\tilde{e}_{i k} \in[0,1]$, $c_{i, r}=\left\|\mathcal{M}_{i}^{r}\right\|$.

After getting the encoded feature vector, we need to use the decoder to map the latent feature distribution to the input $\xi$. The design of the Encoder determines the ability of feature recovery, while simple decoders (such as multilayer perceptrons) are less capable and cannot recover high-level semantic information. In this paper, we choose the Graph Attention Network (GAT) with stronger decoding ability as the decoder of SMGAE, which can utilize the surrounding neighbor information to recover the input features instead of just relying on the nodes themselves.

In the process of coding and decoding, we do not use the mean square error (MSE). Because it is easily affected by the vector dimension and norm, but uses the cosine similarity error, which is more stable in the training process and guide the optimization direction of the model gradient. The formula is defined as follows:
\begin{equation}
	\begin{aligned}
		\cos \left(\xi_{i}, Z_{i}\right) &=\frac{\sum_{m=1}^{N}\left(\xi_{i} \cdot Z_{i}\right)}{\sqrt{\sum_{m=1}^{N}\left(\xi_{i}\right)^{2}} \cdot \sqrt{\sum_{m=1}^{N}\left(Z_{i}\right)^{2}}}+\lambda\|W\|_{F}^{2} \\
		&=\sum_{m=1}^{N} \frac{\xi_{i} \cdot\left(\widetilde{D}^{-\frac{1}{2}} \tilde{A} \widetilde{D}^{-\frac{1}{2}} \xi_{i} W\right)}{\sqrt{\left(\xi_{i}\right)^{2} \cdot\left(\widetilde{D}^{-\frac{1}{2}} \tilde{A} \widetilde{D}^{-\frac{1}{2}} \xi_{i} W\right)^{2}}}+\lambda\|W\|_{F}^{2}
	\end{aligned}
\end{equation}
where, $Z_{i}=\widetilde{D}^{-\frac{1}{2}} \tilde{A} \widetilde{D}^{-\frac{1}{2}} \xi_{i} W$ is the feature vector decoded by the graph neural network. $\lambda$ is a hyperparameter, and $\|W\|_{F}^{2}$ is the weight decay coefficient of the model, which is used to improve the robustness of the model.

In this paper, we define $\hat{A}=\widetilde{D}^{-\frac{1}{2}} \tilde{A} \widetilde{D}^{-\frac{1}{2}}$, and the loss function becomes:
\begin{equation}
	\begin{aligned}
		\mathcal{L}(W)^{node} &=\operatorname{tr}\left(\frac{\xi \cdot(\hat{A} \xi W)}{\sqrt{\xi^{T} \xi} \cdot \sqrt{\left[(\hat{A} \xi W)^{T}(\hat{A} \xi W)\right]}}\right)+\lambda\|W\|_{F}^{2} \\
		&=\operatorname{tr}\left(\frac{\xi \hat{A} \xi W}{\sqrt{\xi^{T} \xi} \sqrt{W^{T} \xi^{T} \hat{A}^{T} \hat{A} \xi W}}\right)+\lambda\|W\|_{F}^{2}
	\end{aligned}
\end{equation}
Where A$tr(\cdot)$ is the trace of the matrix. Then we can get the first-order partial derivative of $\mathcal{L}$ to $W$, and set the value of the first-order partial derivative to 0 to obtain the optimal network parameter $W$. The formula is defined as follows:
\begin{equation}
	\begin{aligned}
		\frac{\partial \mathcal{L}}{\partial W}=& \frac{\xi \hat{A} \sqrt{\xi \xi^{T}} \sqrt{\xi W^{T} \xi^{T} \hat{A}^{T} \hat{A} \xi W}}{\xi \xi^{T} \xi W^{T} \xi^{T} \hat{A}^{T} \hat{A} \xi W} \\
		&-\frac{\xi \hat{A} \xi W \sqrt{\xi^{T} \xi} \frac{\sqrt{W^{T} \xi^{T} \hat{A}^{T} \hat{A}} \xi}{\sqrt{W}}}{\xi \xi^{T} \xi W^{T} \xi^{T} \hat{A}^{T} \hat{A} \xi W}+2 \lambda W \\
		&=0
	\end{aligned}
\end{equation}

For the reconstruction of the edge structure, we will use the contrastive loss of positive and negative samples to optimize, the formula is defined as follows:
\begin{equation}
	\mathcal{L}_{i}^{e d g e}=-\sum_{\varkappa^{+} \in e_{i}^{[M]}} \log \frac{\exp \left(D_{i}^{e d g e}, D_{\varkappa^{+}}^{e d g e}\right)}{\sum_{j \in M_{i}^{-} \cup\left\{\varkappa^{+}\right\}} \exp \left(D_{i}^{e d g e}, D_{\varkappa}^{e d g e}\right)}
\end{equation}
where, $\varkappa^+$ represents the masked edge, and $D_i^{edge}$ represents the probability of the edge belonging to the $i$-th node.

\subsubsection{Weighted Relation-aware Multi-subgraph Information Aggregation}
To better fuse the multiple information between relations and capture the correlation between different relations, we design a Multiple Information Transformer (MIT) to aggregate the interactive information between different relations through multiple information fusion. After modeling the information aggregation of multiple subgraphs, the sentiment classification effect of DER-GCN will be more credible.

\begin{figure}
	\centering
	\includegraphics[width=0.95\linewidth]{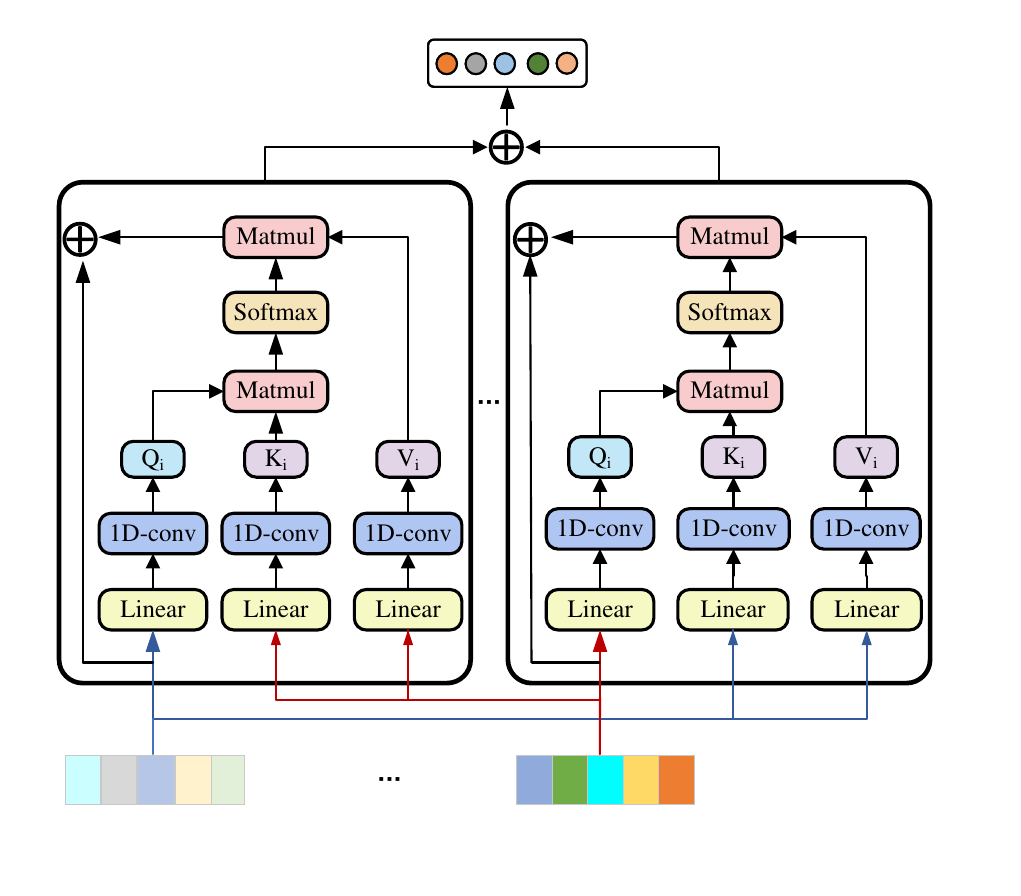}
	\caption{The Multiple Information Transformer (MIT) consists of multiple Transformer modules, each containing multiple linear layers, 1D-Conv, and softmax layers. MIT captures the underlying joint distribution between different relations by transferring information.}
	\label{fig:transformer}
\end{figure}

As shown in the Fig. \ref{fig:transformer}, MIT is composed of Transformers with multiple cross branches, and the interactive information between different relations will be bidirectionally transmitted in MIT. Specifically, we first input the feature vectors obtained after masked graph autoencoder learning into three fully connected layers and 1D convolutional layers, respectively, to obtain vectors $Q$, $K$, $V$. The formula is defined as follows:
\begin{equation}
	\begin{aligned}
		&{\left[Q_{i}^{1}, Q_{i}^{2}, \ldots, Q_{i}^{N}\right]=\operatorname{Conv}\left(\left[I_{i}^{1}, I_{i}^{2}, \ldots, I_{i}^{N}\right] W_{Q}^{\mathbb{R}^{d_{I}}}\right)} \\
		&{\left[K_{i}^{1}, K_{i}^{2}, \ldots, K_{i}^{N}\right]=\operatorname{Conv}\left(\left[I_{i}^{1}, I_{i}^{2}, \ldots, I_{i}^{N}\right] W_{K}^{\mathbb{R}^{d_{I}}}\right)} \\
		&{\left[V_{i}^{1}, V_{i}^{2}, \ldots, V_{i}^{N}\right]=\operatorname{Conv}\left(\left[I_{i}^{1}, I_{i}^{2}, \ldots, I_{i}^{N}\right] W_{V}^{\mathbb{R}^{d_{I}}}\right)}
	\end{aligned}
\end{equation}
where, $W_Q^{\mathbb{R}^{d_I}}$, $W_K^{\mathbb{R}^{d_I}}$, $W_V^{\mathbb{R}^{d_I}}$ are the learnable network parameters in the fully connected layer, and Conv is a one-dimensional convolution operation. Next, we use the softmax function to obtain the attention scores for feature vectors composed of different relations as follows:
\begin{equation}
	\begin{aligned}
		&{\left[\text { att }_{\text {score }}^{1}, \operatorname{att}_{\text {score }}^{2}, \ldots, a t t_{\text {score }}^{N}\right]_{i}} \\
		&=\operatorname{softmax}\left(\frac{\left[Q_{i}^{1}, Q_{i}^{2}, \ldots, Q_{i}^{N}\right]\left[K_{i}^{1}, K_{i}^{2}, \ldots, K_{i}^{N}\right]^{T}}{\varepsilon}\right)
	\end{aligned}
\end{equation}
Where $\varepsilon$ is the dimension of the feature vector $Q$. $T$ represents the transposition of the matrix. Finally, we perform information fusion across relations by the following formula:
\begin{equation}
	\begin{aligned}
		\hat{I}_{i}^{\vartheta}=& I_{i}^{\vartheta}+\left[\text { att }_{\text {score }}^{1}, a t t_{\text {score }}^{\vartheta-1}, \ldots, a t t_{\text {score }}^{\vartheta+1}, \text { att }_{\text {score }}^{N}\right]_{i} \\
		& {\left[V_{i}^{1}, V_{i}^{\vartheta-1}, \ldots, V_{i}^{\vartheta+1}, V_{i}^{N}\right] }
	\end{aligned}
\end{equation}
Where, $\vartheta$ represents the $\vartheta$-th relation. After cross-relational information fusion, we can obtain multi-relational fusion vectors containing rich semantic information.

\subsubsection{Balanced Sampling Strategy-based Contrastive Learning Mechanism}
The number of emotions in each category in the MERC task is quite different. If the cross-entropy loss function is used to guide the learning process of the model, it will cause the model to have a serious overfitting effect on the minority category of emotions. Inspired by contrastive learning, it can learn discriminative boundary information for instances between classes. Therefore, it effectively alleviates the long-tail problem in MERC.

Based on the above research, we introduce a triplet loss function in the process of model training to solve the problem of class distribution imbalance. In addition, we also add a global cross-entropy loss to preserve as much graph structure information as possible.

For each utterance $m_i$, we sample its positive samples $m_i^+$ and negative samples $m_i^-$ to get the triplet loss value of the model, which narrows the gap between positive samples and actual samples. It can widen the gap between negative samples and actual samples. The formula is defined as follows:
\begin{equation}
	\begin{aligned}
		&\mathcal{L}_{E}\\=
		&\sum_{\left(\varkappa_{m_{i}}, \varkappa_{m_{i}}^+, 	\varkappa_{m_{i}}^-\right) \in S} \max \left\{E\left(\varkappa_{m_{i}}, \varkappa_{m_{i}}^{+}\right)-E\left(\varkappa_{m_{i}}, \varkappa_{m_{i}}^-\right)+b, 0\right\}
	\end{aligned}
\end{equation}
Where, $E(,)$ is used to calculate the Euclidean distance between two feature vectors. $b$ is a hyperparameter of the model that measures the distance between samples.

We also construct a global cross-entropy loss to preserve the information of similar structures better. The formula is defined as follows:
\begin{equation}
	\mathcal{L}_{C}=-\frac{1}{\sum_{m=1}^{\sigma} \mathcal{L}_{i}} \sum_{m=1}^{\sigma} \sum_{n=1}^{\gamma_{n}} \sum_{k=1}^{\lambda} y_{m, k}^{n} \log _{2}\left(\hat{y}_{m, k}^{n}\right)
\end{equation}
where $\theta$ is the total number of dialogues in the benchmark dataset, $\gamma_n$ represents the number of utterances in the $n$-th dialogue, and $\lambda$ is the total number of sentiment categories.
\subsubsection{Emotion Classification}
The emotional features $E_f$ obtained after going through the graph convolutional neural network is sent to a linear layer with residual connections, and then goes through a layer of softmax layer to obtain the probability distribution $P$ of emotional labels: the formula is defined as follows:
\begin{equation}
	\begin{gathered}
		\alpha=E_{f}+\operatorname{ReLU}\left(E_{f} W_{f}+b_{f}\right) \\
		P=\operatorname{softmax}\left(\alpha W_{\alpha}+b_{\alpha}\right)
	\end{gathered}
\end{equation}
where, $W_f \in R^{d_f\times d_f}$, $b_f\in \mathbb{R}^{d_f}$, $W_\alpha \in \mathbb{R}^{d_f\times d_\lambda }$, $b_\alpha \in \mathbb{R}^\lambda$ are parameters that can be learned in the model.

We get the sentiment label with the maximum probability through the argmax function:
\begin{equation}
	\widehat{y}=\operatorname{argmax}(P)
\end{equation}
where $\hat{y}$ represents the sentiment label predicted by the model.


\section{Experiments}
\subsection{Benchmark Dataset Used}

The IEMOCAP \cite{busso2008iemocap} and MELD \cite{poria-etal-2019-meld} benchmark datasets are two multimodal dialogue sentiment datasets that researchers widely use to evaluate the effectiveness of their models.

The Interactive Emotional Dyadic Motion Capture Database (IEMOCAP) is a multimodal emotion recognition dataset. The IEMOCAP dataset contains three modalities of the speaker's video, audio, and dialogue text. The dataset contains 5 actors and 5 actresses, and each dialogue scene has a dialogue between an actor and an actress. The labels of these conversations are all manually annotated, and at least three experts in the emotion domain are assigned to each conversation.

The Multi-modal EmotionLines Dataset (MELD) is a popular multi-modal benchmark dataset in the MDER domain, consisting of multiple dialogue clips from the TV series friends. The total video and audio duration of MELD is approximately 13.7 hours, and each video clip contains multiple speakers. The labels of these conversations are all manually annotated, and at least five experts in the emotion domain are assigned to each conversation.

\subsection{Evaluation Metrics}
In this section, we illustrate the evaluation metrics used to verify the effectiveness of the model proposed in this paper. This paper uses the following four evaluation metrics: 1) accuracy; 2) f1; 3) weight accuracy (WA); 4) weight f1 (WF1). Due to the serious data imbalance problem in the IEMOCAP and MELD benchmark datasets, we will mainly use WA and WF1 as our main evaluation metrics. 

\subsection{Baseline Models}
To verify the effectiveness of our model on the IEMOCAP and MELD benchmark datasets, we conduct comparative experiments with twelve state-of-the-art deep learning-based algorithms, including one traditional CNN algorithm (i.e., TextCNN \cite{kim-2014-convolutional}), four RNN algorithms (i.e., bc-LSTM \cite{poria-etal-2017-context}, DialogueRNN \cite{Majumder_Poria_Hazarika_Mihalcea_Gelbukh_Cambria_2019}, CMN \cite{hazarika2018conversational}, and A-DMN \cite{9128015}), three GNN algorithms (i.e., DialogueGCN \cite{ghosal-etal-2019-dialoguegcn}, RGAT \cite{ishiwatari2020relation}, LR-GCN \cite{9556142}), one feature fusion algorithm (i.e., LFM \cite{liu-etal-2018-efficient-low}), and three pre-trained algorithms (CoMPM \cite{lee2022compm}, EmoBERTa \cite{kim2021emoberta}, and COGMEN \cite{joshi2022cogmen}).

\subsection{Implementation Details}

To effectively evaluate the experimental effect of the algorithm in this paper, we divide the dataset into the training set, validation set, and test set, and the ratio of the test set to validation set is 4:1. The test set is used to evaluate the training effect of the model. The validation set is used to fine-tune the model parameters. All experiments in this paper are carried out on a server with Ubuntu 18.04 operating system, hardware model Nvidia RTX 3090, and a video memory capacity of 24G. Our Python version is 3.7, and the Pytorch version is 1.8.1. We choose the Adam optimization algorithm \cite{kingma2015adam} for gradient updates in the updating process of model parameters. A total of 60 iterations of training are performed in this experiments, and the batch size during each iteration is 32. We set the learning rate to 0.0003, dropout to 0.25, and L2 regularization term to 0.0001.

\section{Results and Discussion}
\begin{table*}[!t]
	\renewcommand\arraystretch{1.5}
	\setlength{\tabcolsep}{15.8pt}
	\caption{Comparison with other baseline models on the IEMOCAP dataset, Acc.=Accuracy, Average(w) = Weighted average.}
	\begin{tabular}{l|ccccccc}
		\hline
		\multirow{3}{*}{Methods} & \multicolumn{7}{c}{IEMOCAP}                                                              \\ \cline{2-8}
		& Happy      & Sad        & Neutral    & Angry      & Excited    & Frustrated & Average(w) \\ \cline{2-8}
		& Acc.  F1   & Acc.  F1   & Acc.  F1   & Acc.  F1   & Acc.  F1   & Acc.  F1   & Acc.  F1   \\ \hline
		TextCNN                      & 27.7  29.8 & 57.1  53.8 & 34.3  40.1 & 61.1  52.4 & 46.1  50.0 & 62.9  55.7 & 48.9  48.1 \\
		bc-LSTM                  & 29.1  34.4 & 57.1  60.8 & 54.1  51.8 & 57.0  56.7 & 51.1  57.9 & {67.1}  58.9 & 55.2  54.9 \\
		CMN                      & 25.0  30.3 & 55.9  62.4 & 52.8  52.3 & 61.7  59.8 & 55.5  60.2 & \textbf{71.1}  60.6 & 56.5  56.1 \\
		LFM              & 25.6  33.1 & 75.1  78.8 & 58.5  59.2 & 64.7  65.2 & {80.2}  71.8 & 61.1  58.9 & 63.4  62.7 \\
		A-DMN              & 43.1  50.6 & 69.4  76.8 & 63.0  62.9 & 63.5  56.5 & \textbf{88.3  77.9} & 53.3  {55.7} & 64.6  64.3 \\
		LR-GCN                & {54.2  55.5} & \textbf{81.6}  79.1 & 59.1  \textbf{63.8} & 69.4  69.0 & 76.3  {74.0} & 68.2  \textbf{68.9} & 68.5  68.3 \\
		DER-GCN                   & \textbf{60.7  58.8} & 75.9  \textbf{79.8} & \textbf{66.5}  61.5 & \textbf{71.3  72.1} & 71.1  73.3 & 66.1  67.8 & \textbf{69.7  69.4} \\ \hline
	\end{tabular}
\end{table*}

\begin{table*}[!t]
	\renewcommand\arraystretch{1.5}
	\caption{Comparison with other baseline models on the MELD dataset, Acc.=Accuracy, Average(w) = Weighted average.}
	\setlength{\tabcolsep}{11.65pt}{
		\begin{tabular}{l|cccccccc}
			\hline
			\multirow{3}{*}{Methods} & \multicolumn{8}{c}{MELD}                                                                            \\ \cline{2-9}
			& Neutral     & Surprise    & Fear     & Sadness    & Joy        & Disgust  & Anger      & Average(w) \\ \cline{2-9}
			& Acc.  F1    & Acc.  F1    & Acc.  F1 & Acc.  F1   & Acc.  F1   & Acc.  F1 & Acc.  F1   & Acc.  F1   \\ \hline
			TextCNN                      & {76.2}  74.9  & 43.3  45.5  & 4.6  3.7 & 18.2  21.1 & 46.1  49.4 & 8.9  8.3 & 35.3  34.5 & 56.3  55.0 \\
			bc-LSTM                  & 78.4   73.8 & 46.8   47.7 & 3.8  5.4 & 22.4  25.1 & 51.6  51.3 & 4.3  5.2 & 36.7  38.4 & 57.5  55.9 \\
			DialogueRNN              & 72.1   73.5 & 54.4  49.4  & 1.6  1.2 & 23.9  23.8 & 52.0  50.7 & 1.5  1.7 & 41.0  41.5 & 56.1  55.9 \\
			A-DMN                &  76.5 78.9       &  \textbf{56.2} 55.3        &   8.2 8.6     &   22.1  24.9       &  59.8 57.4      &   1.2 3.4     &   41.3  40.9      & 61.5  60.4 \\
			LR-GCN                   & \textbf{81.5}   \textbf{80.8} & 55.4  \textbf{57.1}  & 0.0  0.0 & 36.3  36.9 & 62.2  \textbf{65.8} & 7.3  11.0 & \textbf{52.6}  54.7 & 65.7  65.6 \\
			DER-GCN                   &76.8 {80.6} & 50.5 51.0 & \textbf{14.8 10.4} & \textbf{56.7 41.5} & \textbf{69.3} 64.3 &\textbf{17.2 10.3} &52.5 \textbf{57.4} &\textbf{66.8 66.1}\\ \hline
	\end{tabular}}
\end{table*}

\subsection{Comparison with Baselines}
To verify the effectiveness of the DER-GCN model proposed in this paper, we have done extensive experiments to compare it with other comparison algorithms. Tables 1 and 2 present the emotion recognition effects of DER-GCN and other comparative algorithms on two popular datasets, respectively.

\textbf{IEMOCAP:} As shown in Table 1, compared with other comparison algorithms, our proposed multimodal dialogue emotion recognition method DER-GCN has the best emotion recognition effect on the IEMOCAP dataset, and the WA and WF1 values are 69.7\% and 69.4\%, respectively. DER-GCN proposes a method for dialogue emotion recognition that comprehensively considers sequential context information, dialogue relations between speakers, and event relations. Among other benchmark models, LR-GCN performs slightly worse than DER-GCN, with WA and WF1 values of  68.5\% and 68.3\%, respectively. We speculate that LR-GCN outperforms other baseline models because it considers both the interaction between speakers and the latent semantic relationship of the dialogue context. However, LR-GCN ignores the event relations in the dialogue, so its emotion recognition effect is lower than that of the model proposed in this paper, DER-GCN. The emotion prediction effect of A-DMN and LFM is much lower than that of DER-GCN, with WA values of 64.6\% and 63.4\%, and WF1 values of 64.3\% and 62.7\%, respectively. It is because they do not model speaker relations and event relations in dialogue although they design a fusion mechanism to obtain complementary multimodal semantic features. The emotion prediction performance of other baseline methods, such as TextCNN, is much worse than that of DER-GCN, because they only model sequential context information, which results in limited semantic information learned by the model.

\textbf{MELD:} As shown in Table 2, the emotion prediction effect of the DER-GCN model on the MELD dataset is better than other comparison algorithms, and the WA and WF1 values are  66.8\% and 66.1\%, respectively. The effect of LR-GCN is second, with WA and WF1 values of 65.7\% and  65.6\%, respectively. The prediction performance of A-DMN is lower than that of DER-GCN and LR-GCN, with WA and WF1 values of 61.5\% and 60.4\%, respectively. Other comparison algorithms perform poorly because they all ignore modeling the relationships between speakers. In addition, compared with other comparison algorithms, DER-GCN has significantly improved the prediction accuracy on the minority class sentiment labels ``fear" and ``disgust". Specifically, the WA and WF1 values of DER-GCN on the ``fear" label are 14.8\% and 10.4\%, respectively, and the prediction effect is improved by about 10\%. The WA and WF1 values of DER-GCN on the ``disgust" label are 17.2\% and 10.3\%, respectively, and the prediction effect is improved by about 10\%. We have guessed that DER-GCN can improve the prediction effect of minority class sentiment. The model adopts a loss optimization strategy based on the contrastive learning mechanism, which can better represent minority class features.

The experimental results show that the event relationship in the dialogue significantly strengthens the model's understanding of the speaker's emotion. In addition, cross-modal feature fusion and loss optimization strategy based on contrastive learning can also enhance the model's emotion classification ability.

\subsection{Analysis of the Experimental Results}
To clarify the feature representation ability of the model on each emotion category, we analyze the distribution of emotion classification of DER-GCN on the test set. Fig. 5 presents the confusion matrix for emotion classification by DER-GCN on the IEMOCAP and MELD datasets.

\begin{figure*}[htbp]
	\centering
	\includegraphics[width=1\linewidth]{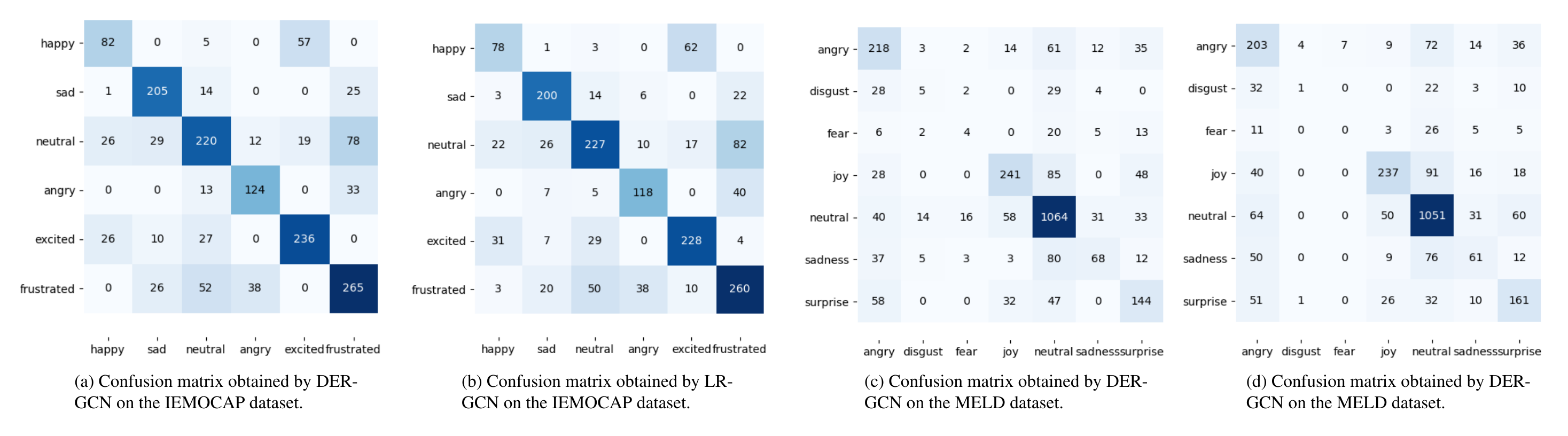}
	\caption{The classification of DER-GCN and LR-GCN on the IEMOCAP and MELD dataset.}
	\label{fig:}
\end{figure*}

On the IEMOCAP dataset, we observe the confusion matrix and find that DER-GCN easily misclassifies ``neutral" sentiment into ``frustrated" and ``sad" sentiment. We believe that this is because there are semantically similar parts between ``neutral" sentiment and ``frustrated" or ``sad" sentiment, which leads to fuzzy class boundaries in the representation of emotional features among different categories learned by DER-GCN. At the same time, we also find that the model incorrectly classified ``frustrated" or ``sad" sentiment as ``neutral" sentiment. In addition, DER-GCN also has mutual misclassification between ``sad" sentiment and ``frustrated" sentiment. There is also overlapping semantic information between ``happy" sentiment and "excited" sentiment, which makes it difficult for DER-GCN to distinguish these emotions. The classification effect of the model in ``sad" or ``angry" sentiment is relatively good. Most of the tested utterances can be correctly classified. For the "excited" sentiment, we find that DER-GCN misclassifies it as the "sad" sentiment. We think this is because speakers usually express their emotions more implicitly and sarcastically when they talk about sensitive topics, and the model cannot capture this semantic information. 

The MELD dataset shows a specific semantic correlation between the ``neutral" sentiment and other types of emotion. Therefore, DER-GCN is prone to misclassify the ``neutral" sentiment as other emotions. The opposite is also true. For the ``surprise" sentiment, DER-GCN incorrectly classifies it into ``joy" and ``anger" sentiment. We guess this is because speakers with ``surprise" sentiments are usually accompanied by ``joy" or "anger" sentiments.

On the one hand, the speaker is stimulated by something wrong to produce surprise-like emotions, which will cause the speaker to feel angry. On the other hand, the speaker is surprised by surprise prepared by others, which will cause the speaker to feel joy. For the ``fear" sentiment, the model is prone to misclassify it as the ``neutral" sentiment. For the ``disgust" sentiment, the number of test utterances correctly classified by DER-GCN is minimal, and the classification results are unreliable. This is because the number of ``disgust" sentiments in the MELD dataset is very small. DER-GCN cannot learn effective semantic information from such a small amount of data. At the same time, this problem also exists in the ``fear" category of emotions. For the ``angry" sentiment, DER-GCN not only misclassifies it as the ``surprise" sentiment but also misclassifies it as the ``sadness" or the ``joy" sentiment. On the one hand, speakers with ``angry" emotions are usually accompanied by ``sadness" sentiments. On the other hand, speakers with an ``angry" sentiment may be more implicit in expressing their emotions. The above two reasons may lead to biases in DER-GCN in understanding the semantics of test utterances.

\subsection{Importance of the Modalities}
To verify the importance of the three modal features of text, video, and audio, we conduct experiments on the IEMOCAP and MELD datasets to compare the performance of unimodal, bimodal, and multimodal features. The experimental results are shown in the Table 3. Due to the problem of data imbalance in the dataset, WF1 comprehensively considers the precision rate and recall rate. So we choose WF1 as our main evaluation metric and WA as our secondary evaluation metric. For the experimental results of the single modality, the values of WA and WF1 of the text modality are higher than the audio and video modality. The values of WA are 63.2\% and 62.8\% on the IEMOCAP and MELD datasets, respectively, and the values of WF1 are 63.8\% and 61.9\% on the IEMOCAP and MELD datasets, respectively, which indicates that the text modal features play the most important role in the emotion recognition of the model. The effect of the audio modality is second, the values of WA are 61.4\% and 62.1\%, respectively, and the values of WF1 are 61.6\% and 61.3\%, respectively. The video modality performs the worst, with values of 57.8\% and 60.5\% for WA, respectively, and with values of 57.1\% and 60.6\% for WF1, respectively, indicating that it is difficult for the model to extract useful emotional features from video features. The experimental results show that the noise introduced by the text features is the least, which will benefit the model in learning the embedded representation of the emotional features.

\begin{table}[htbp]
	\renewcommand\arraystretch{1.5}
	\setlength{\tabcolsep}{5.6mm}{
		\caption{The effect of DER-GCN on two datasets using unimodal features and multimodal features, respectively. T, V, and A represent text, video, and audio modality features.}
		\begin{tabular}{c|cccc}
			\hline
			\multirow{2}{*}{Modality} & \multicolumn{2}{c}{IEMOCAP} & \multicolumn{2}{c}{MELD} \\ \cline{2-5} 
			& WA       & WF1              & WA         & WF1         \\ \hline
			T                         & 63.2        & 63.8       & 62.8          & 61.9           \\
			A                         & 61.4        & 61.6                & 62.1          & 61.3           \\
			V                       & 57.8        & 57.1                & 60.5          & 60.6
			\\
			T+A                       & 65.8        & 64.7                & 63.8          & 62.6           \\
			T+V                       & 64.4        & 64.0                & 63.1          & 63.4           \\
			V+A                       & 61.2        & 60.9                & 60.3          & 59.8           \\
			T+A+V                     & \textbf{69.7}        & \textbf{69.4}                & \textbf{66.8}          & \textbf{66.1}           \\ \hline
	\end{tabular}}
\end{table}

The experimental results of bi-modality are better than single-modality. The WA value is improved by 0.2\% to 8\%, and the WF1 value is improved by 0.7\% to 7\%. It indicates that emotional features are not only related to contextual information but also changes in sound signals in audio and facial expressions in video. The bimodal feature combines two different unimodal features, which can effectively improve the emotion prediction effect of the model. Furthermore, the bimodal features fused with text and audio performed the best emotion prediction, with values of 65.8\% and 63.8\% for WA, respectively, and with values of 64.7\% and 62.6\% for WF1, respectively. The emotion prediction effect of bimodal features fused by text and video is second, with the values of 64.4\% and 63.1\% for WA, respectively, and with the values of 64.0\% and 63.4\% for WA, respectively. The bimodal features fused with audio and video have the worst emotion prediction performance, with values of 61.2\% and 60.3\% for WA, respectively, and with values of 60.9\% and 59.8\% for WF1, respectively.

After the fusion of three modal features of text, video, and audio, the multi-modal features have the best emotion prediction performance. It is better than the performance of single-modal and bi-modal features, which indicates that the model not only utilizes the semantic information of the dialogue context. It also utilizes video and audio features to enhance the representation ability of the emotional feature vectors.

\subsection{Effectiveness of Cross-modal Feature Fusion}
In this section, to verify the effectiveness of the cross-modal feature fusion method proposed in this paper, we compare it with other three fusion methods,. i.e., add and concatenation operation, and tensor fusion network (TFN).

%
%

The experimental results are shown in Table 4. Compared with other multi-modal feature fusion methods, the cross-modal feature fusion method proposed in this paper has achieved the best experimental results, and the values of WA are 69.7\% and 66.8\%, respectively, and the values of WF1 are 69.4\% and 66.1\%, respectively. Specifically, compared with the Add method, the WA value of the cross-modal feature fusion method is improved by 3.9\% to 4.5\%, and the WF1 value is improved by 3.7\% to 4.6\%. We think the Add method cannot capture the complementary semantic information between different modalities. The cross-modal feature fusion method can extract the most relevant semantic information with emotional features through the attention mechanism, thereby improving the emotion recognition effect of the model. At the same time, compared with the Concatenate method, the WA value of the cross-modal feature fusion method is improved by 4.3\% to 5.1\%, and the WF1 value is improved by 4.5\% to 5.3\%. The reason is that the feature vector dimensions of the text, video, and audio modalities are high, leading to the combinational explosion of multimodal embedding representations generated by feature concatenation. Different from the Concatenate method, the cross-modal feature fusion method can achieve efficient feature dimensionality reduction while capturing rich semantic information. In addition, compared with the Tensor Fusion method, the WA value of the cross-modal feature fusion method is improved by 3\% to 3.1\%, and the WF1 value is improved by 2.6\% to 3.8\%. It is because the Tensor Fusion method needs to use tensors for feature representation, which introduces much computational consumption and reduces emotion recognition accuracy. The above experimental results demonstrate the effectiveness of the cross-modal feature fusion method proposed in this paper.

\begin{table}[htbp]
	\renewcommand\arraystretch{1.5}
	\setlength{\tabcolsep}{3.4mm}{
		\caption{Emotion recognition effects of different multimodal feature fusion methods on IEMOCAP and MELD datasets. We use three modal text, video, and audio features for each method.}
		\begin{tabular}{c|cccc}
			\hline
			\multirow{2}{*}{Methods} & \multicolumn{2}{c}{IEMOCAP} & \multicolumn{2}{c}{MELD} \\ \cline{2-5} 
			& WA       & WF1              & WA         & WF1         \\ \hline
			Add                      & 65.2        & 64.8       & 62.9          & 62.4           \\
			Concatenate              & 64.6        & 64.1               & 62.5          & 61.6           \\
			Tensor Fusion            & 66.7        & 65.6                & 63.7          & 63.5           \\
			Cross-modal Fusion(Ours) & \textbf{69.7}        & \textbf{69.4}                & \textbf{66.8}          & \textbf{66.1}           \\ \hline
	\end{tabular}}
\end{table}

\begin{figure}
	\centering
	\includegraphics[width=1\linewidth]{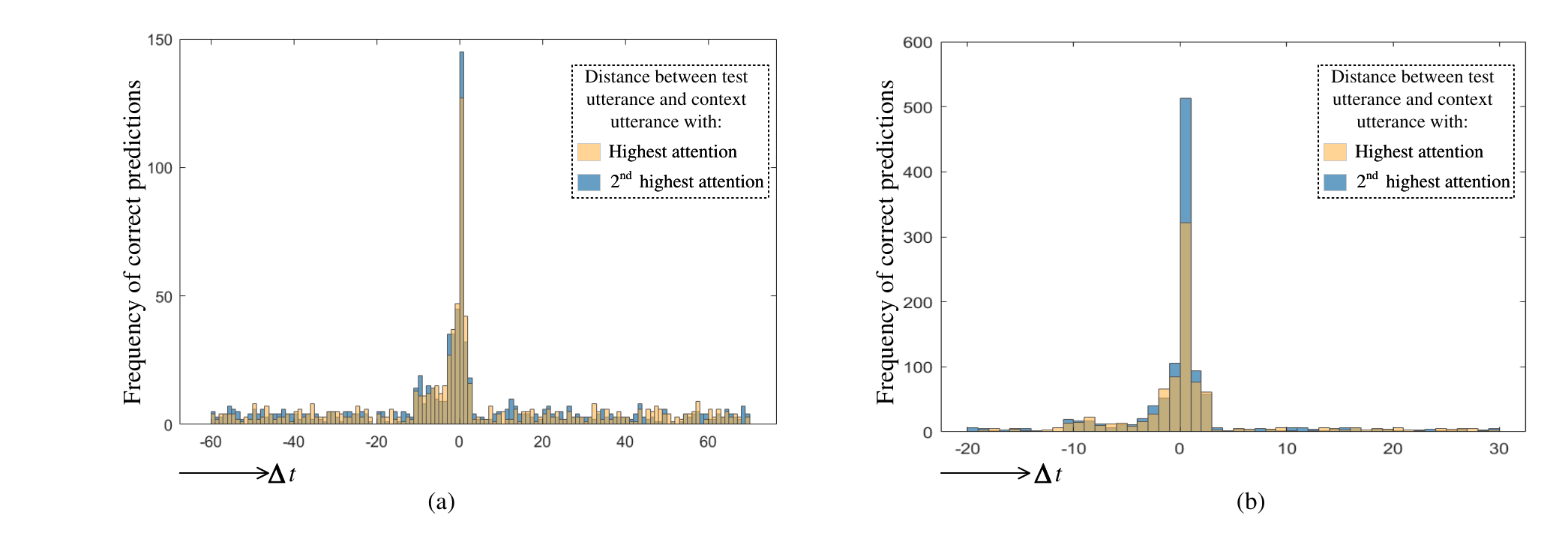}
	\caption{Histograms of attention scores for IEMOCAP and MELD datasets.}
	\label{fig:attention}
\end{figure}

\subsection{Effectiveness of Bi-GRU}
To verify the effectiveness of Bi-GRU for contextual semantic information extraction, we use three methods for comparative experiments. The experimental results are shown in Table 5.

\begin{itemize}
	\item Without contextual modeling: This method does not use any contextual information modeling method for emotion recognition. Specifically, we replace the GRU layers with linear layers.
	
	\item Unidirectional GRU (Uni-GRU): Instead of modeling context information, we use a unidirectional GRU to extract contextual semantic information, which can memorize utterance information before the current moment.
	
	\item Bidirectional GRU (Bi-GRU): Different from the above methods, we use Bi-GRU to model two opposite contextual utterances, which contain richer contextual information.
\end{itemize}

\begin{figure*}[htbp]
	\centering
	\subfloat[Raw data distribution]{
		\begin{minipage}[t]{0.24\linewidth}
			\centering
			\includegraphics[width=1.8in]{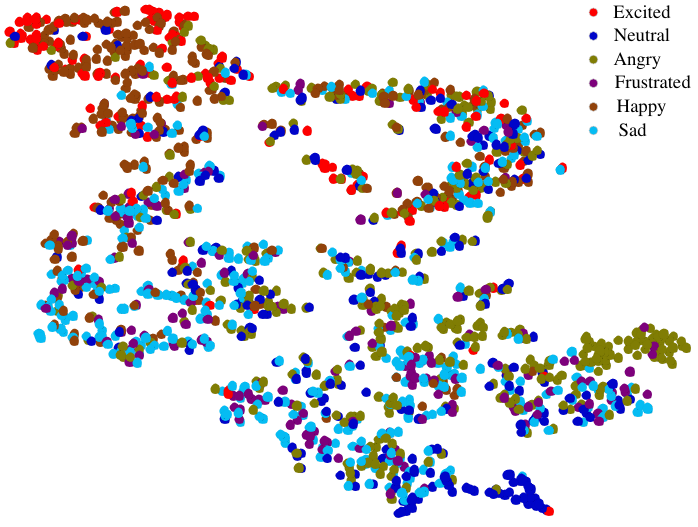}
		\end{minipage}%
	}%
	\subfloat[Learned by bc-LSTM]{
		\begin{minipage}[t]{0.24\linewidth}
			\centering
			\includegraphics[width=1.8in]{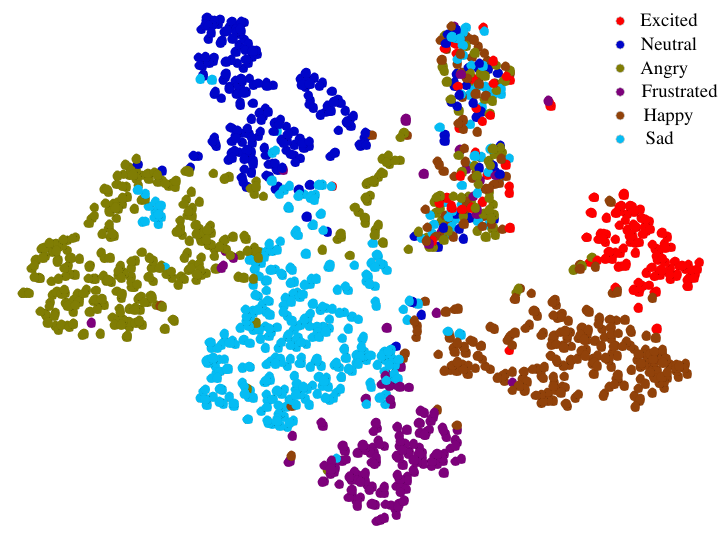}
		\end{minipage}%
	}%
	\subfloat[Learned by DialogueGCN]{
		\begin{minipage}[t]{0.24\linewidth}
			\centering
			\includegraphics[width=1.8in]{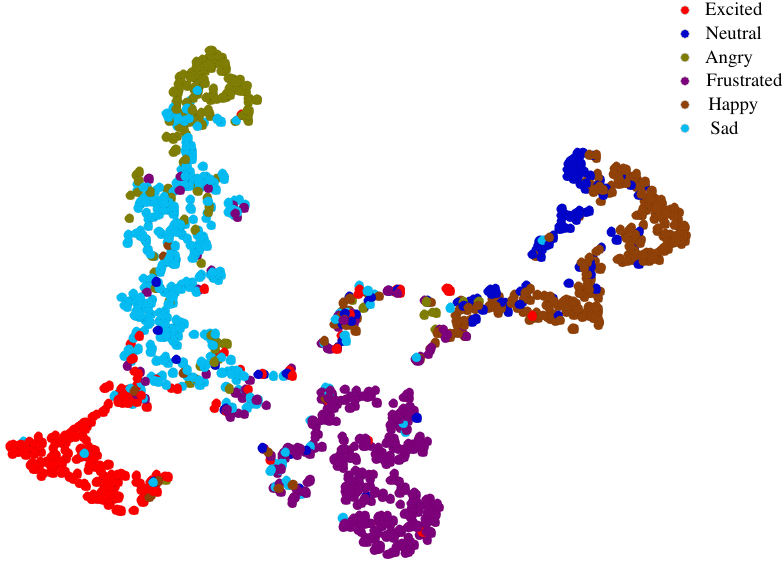}
		\end{minipage}%
	}%
	\subfloat[Learned by DER-GCN]{
		\begin{minipage}[t]{0.24\linewidth}
			\centering
			\includegraphics[width=1.8in]{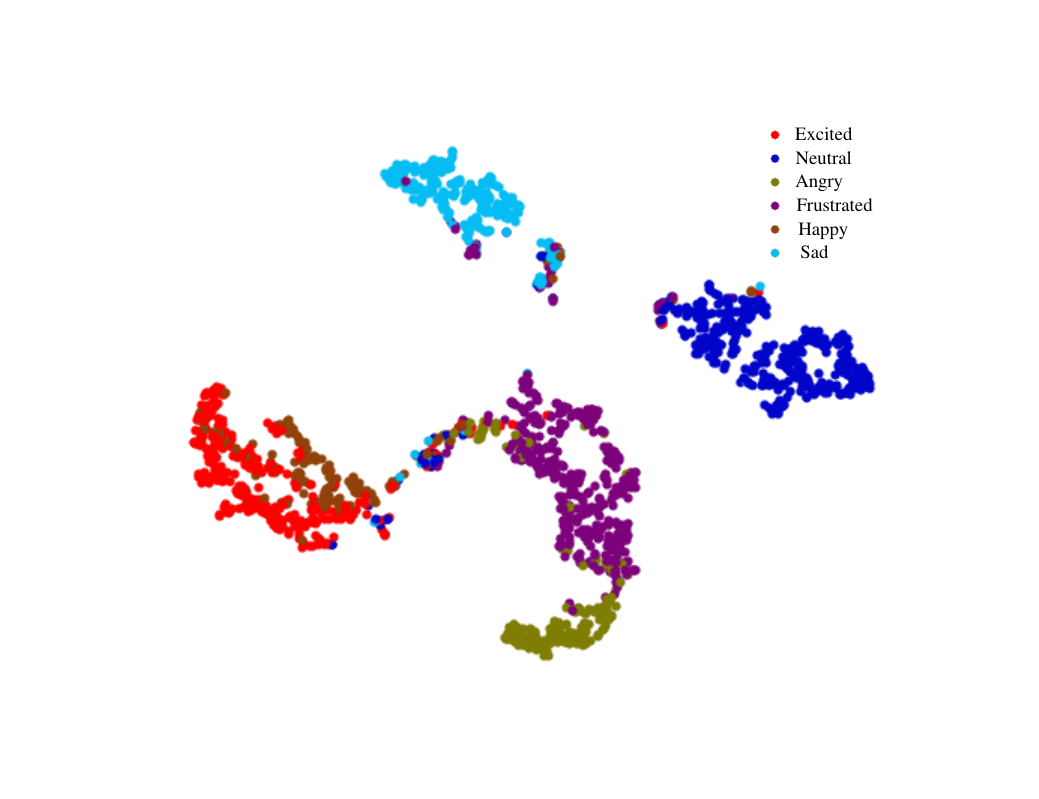}
		\end{minipage}%
	}%
	\centering
	\caption{Visualizing feature embeddings for the multimodal sentiment on the IEMOCAP benchmark dataset. Each dot represents an utterance, and its color represents an emotion.}
	
\end{figure*}

Among the three contrasting methods, we find that the emotion recognition method that does not model contextual semantic information works the worst, with WA values of 62.3\% and 60.1\% on IEMOCAP and MELD datasets, respectively, and with WF1 values of 61.7\% and 61.6\%, indicating the necessity of contextual semantic information modeling. The Uni-GRU method outperforms methods that do not model contextual semantic information, with values of 67.1\% and 63.4\% for WA , respectively, with values of 66.2\% and 63.0\% for WF1 , respectively. Bi-GRU performs the best for emotion recognition, with WA values of 69.7\% and 66.8\%, respectively, and with WF1 values of 69.4\% and 66.1\%. Compared with the other two methods, the WA value is increased by 2.6\% to 7.4\%, and the WF1 value is increased by 3.1\% to 7.7\%. Therefore, the experimental results show that the emotional information of the current moment is related to both historical discourse and future discourse.

\begin{table}[htbp]
	\renewcommand\arraystretch{1.5}
	\caption{Different context modeling methods on the two datasets. All methods have experimented with multimodal features.}
	\setlength{\tabcolsep}{8.6pt}{
		\begin{tabular}{c|cccc}
			\hline
			\multirow{2}{*}{Methods}    & \multicolumn{2}{c}{IEMOCAP} & \multicolumn{2}{c}{MELD} \\ \cline{2-5} 
			& WA       & WF1              & WA         & WF1         \\ \hline
			Without contextual modeling & 62.3        & 61.7     & 60.1          & 61.6          \\
			Uni-GRU                     & 67.1        & 66.2                & 63.4          & 63.0           \\
			Bi-GRU(Ours)                & \textbf{69.7}        & \textbf{69.4}                & \textbf{66.8}          & \textbf{66.1}          \\ \hline
	\end{tabular}}
\end{table}

\subsection{Discussion on Contextual Distances}

We analyze test utterances correctly classified by DER-GCN on the IEMOCAP and MELD datasets. As shown in Fig. 6, we present the distribution of distances between the current test utterance, and the second-highest attended utterance in the context. The utterances with correct sentiment classification have the strongest context-dependence on the current moment. Furthermore, as the distance increases, the dependence of the current utterance on the distant context decreases. However, we found that a significant portion of utterances will focus on contextual utterances with distances between 20 and 30, which indicates the need for contextual semantic information modeling. Therefore, the modeling dialogue texts' long-distance dependencies can improve the model's emotion recognition accuracy.

\subsection{Visualization}

To more intuitively display the emotional feature vectors in high-dimensional space, we use the t-SNE \cite{JMLR:v9:vandermaaten08a} method to reduce the dimensionality of the original emotional features in the IEMOCAP dataset and the emotional features obtained after model learning and obtain a two-dimensional spatial distribution.

As shown in Fig. 7(a), we can see that the original emotional feature distribution of IEMOCAP is very scattered, and there are many overlapping parts between each emotion category, which will cause the model to fail to classify each emotion type correctly. As shown in Fig. 7(b), we find that the feature representations learned by bc-LSTM have ambiguous class boundary information, and the overlap between each emotion category is greatly reduced, which leads to a better feature distribution than the original sentiment feature distribution. It is because bc-LSTM can utilize the contextual information of the utterance to enhance the feature representation ability of each emotion. Therefore, it makes the feature vector more discriminative in the spatial distribution. However, bc-LSTM does not consider the interaction between speakers, and the semantic information learned by the model is insufficient, which leads to limited sentiment classification ability of the learned class boundaries. As shown in Fig. 7(c), the feature representations learned by DialogueGCN have clearer class boundary information than bc-LSTM, which leads to the stronger sentiment classification ability of the model. We believe this is because DialogueGCN considers the semantic information of the extracted sequential context and models the dialogue relationship between speakers. Embedding representations learned by DialogueGCN have richer semantic information, which leads to different emotion categories being more discriminative in spatial distribution. However, DialogueGCN can still not distinguish on some test utterances because the speaker expresses his emotions implicitly and vaguely when talking about specific events, while the model cannot capture this semantic information. As shown in Fig. 7(d), the emotion distribution of the feature representation obtained after learning by DER-GCN works best. DER-GCN comprehensively models long-range context dependencies, dialogue relationships between speakers, and event relationships in dialogue, which enables a speaker's emotion to be reinforced by other speakers' emotions. Therefore, DER-GCN obtains clearer class boundary information than bc-LSTM and DialogueGCN, which makes the model's sentiment classification the best.

\section{Conclusions and Future Work}

This paper proposes the Dialogue and Event Relation-Aware Graph Convolutional Neural Networks (DER-GCN) model, which enables multimodal emotion recognition for multiple dialogue relations. In order to capture the potential semantic information related to the dialogue topic during the dialogue process, we use an event extraction method to extract the main events in the dialogue. In order to obtain better node embedding representation, we design a graph autoencoder based on node and edge masking mechanism, which reconstructs the original graph's topological structure and features vectors through self-supervised learning. We introduce a sampling strategy based on contrastive learning to alleviate the data imbalance problem. DER-GCN is used to learn optimal network parameters in the multimodal emotion recognition task. On the IEMOCAP and MELD benchmark datasets, DER-GCN has greatly improved the effect of emotion recognition compared with other comparison algorithms.

In future research work, we will consider and adopt self-supervised learning algorithms for multimodal emotion recognition, thereby adding large-scale unlabeled dialogue data to our model to improve the generalization ability of the model. Furthermore, we will also consider how to extend our model to other multimodal recognition tasks.

\bibliographystyle{IEEEtran}
\bibliography{refs}


%

%

\ifCLASSOPTIONcompsoc
  \section*{Acknowledgments}
\else
  \section*{Acknowledgment}
\fi

This work is supported by National Natural Science Foundation of China (Grant No. 69189338), Excellent Young Scholars of Hunan Province of China (Grant No. 20B625, No. 18B196), Changsha Natural Science Foundation (Grant No. kq2202294), and program of Research on Local Community Structure Detection Algorithms in Complex Networks (Grant No. 2020YJ009).

\ifCLASSOPTIONcaptionsoff
  \newpage
\fi

\end{document}